\documentclass{ieeetrans} % For LaTeX2e
\usepackage{times}
\usepackage{graphicx}
\usepackage{booktabs}
\usepackage{amsmath}
\usepackage{multirow}

\DeclareMathOperator*{\minimize}{minimize}

\makeatletter
\newcommand{\leqnos}{\tagsleft@true\let\veqno\@@leqno}
\newcommand{\reqnos}{\tagsleft@false\let\veqno\@@eqno}
\reqnos
\makeatother

\usepackage{amsmath,amsfonts,bm}

\def\eqref#1{equation~\ref{#1}}
\def\1{\bm{1}}

\DeclareMathAlphabet{\mathsfit}{\encodingdefault}{\sfdefault}{m}{sl}
\SetMathAlphabet{\mathsfit}{bold}{\encodingdefault}{\sfdefault}{bx}{n}

\newcommand{\R}{\mathbb{R}}

\usepackage{hyperref}
\usepackage{url}

\makeatletter
\let\r@eqnnum\@eqnnum
\input{leqno.clo}
\let\l@eqnnum\@eqnnum
\reqnos
\makeatother
\title{DeepTrax: Embedding Graphs of Financial Transactions}

\author{C. Bayan Bruss, Anish Khazane, Jonathan Rider, Richard Serpe, Antonia Gogoglou, Keegan E. Hines  \\
Capital One\\
McLean, VA 22102, USA \\
}

\begin{document}

\maketitle

\begin{abstract}
Financial transactions can be considered edges in a heterogeneous graph between entities sending money and entities receiving money.  For financial institutions, such a graph is likely large (with millions or billions of edges) while also sparsely connected. It becomes challenging to apply machine learning to such large and sparse graphs.  Graph representation learning seeks to embed the nodes of a graph into a euclidean vector space such that graph topological properties are preserved after the transformation. In this paper, we present a novel application of representation learning to bipartite graphs of credit card transactions in order to learn embeddings of account and merchant entities. Our framework is inspired by popular approaches in graph embeddings and is trained on two internal transaction datasets. This approach yields highly effective embeddings, as quantified by link prediction AUC and F1 score. Further, the resulting entity vectors retain intuitive semantic similarity that is explored through visualizations and other qualitative analyses. Finally, we show how these embeddings can be used as features in downstream machine learning business applications such as fraud detection.
\end{abstract}

\section{Introduction}

Financial transactions between merchants, customers, lenders, and banks present a rich view of the economic activity within a market. It can be useful to consider this type of data as a heterogeneous graph of market participants which are connected by edges (transactions). This is a particularly useful formulation for tackling critical business problems like credit risk modeling, fraud detection, and money laundering detection. However, such a graph will be very high dimensional (with tens or hundreds of millions of vertices) and very sparse (with each vertex interacting with a fraction of the other vertices), thus limiting the graph's utility for common machine learning tasks. 

In recent years, graph embeddings techniques have grown in popularity as a means for learning latent representations of vertices on large-scale networks. Certain techniques like Graph Convolutional Networks (GCNs), DeepWalk, and node2vec attempt to encode topological structure from graphs into dense representations such that nodes with high levels of neighborhood overlap are co-located in the embedding space. This is commonly referred to as geometric similarity, which captures both graphical substructure as well as similarity among any ancillary features - e.g merchant type in the context of financial transactions - that belong to any particular vertex. Embeddings produced by the aforementioned techniques can also serve as useful features for downstream supervised tasks.

In this work, we present a novel application of graph embedding techniques to problems in financial services. In particular, we focus on large-scale datasets of credit card transactions which define an implicit bipartite graph between account-holders and merchants.\footnote{All trademarks referenced herein for illustrative and education purposes only and are owned by their respective owners.} We demonstrate that embedding these transactions can lead to representations which encode economic properties such as spending patterns, geography, and merchant industry/category. In Section 2, we briefly explore current literature on representation learning for large-scale graph networks. We then formally present our own method in Section 3, and quantitatively and qualitatively evaluate our results on multiple financial transaction datasets in Section 4. We conclude  by demonstrating how these embeddings can benefit downstream tasks such as fraud detection. 

\section{Related Work}

\subsection{Types of Graph Embedding Techniques}
 
One way to group graph embedding techniques is based on the type of input they can incorporate. Inputs can be {\it homogeneous} where all nodes are of the same type, {\it heterogeneous} with multiple types of nodes and {\it auxiliary information} graphs that contain node, edge or neighborhood features.  In homogeneous graphs, the challenge is to encode the neighborhood topology of the nodes in a computationally feasible manner \cite{DBLP:conf/aaai/ZhangLZ18, DBLP:conf/aaai/ZhouLLLG17}. The latent vectors are expected to preserve different orders of node proximity (e.g. \cite{DBLP:journals/corr/PerozziAS14}) and different ranges of structural identity (e.g. \cite{DBLP:conf/kdd/RibeiroSF17}, \cite{DBLP:conf/nips/Abu-El-HaijaPAA18}). Therefore, the rich contextual information they carry makes node embeddings useful for multiple unsupervised learning tasks such as predicting missing links \cite{DBLP:journals/corr/GroverL16} as well as recommendation and ranking of the most relevant nodes \cite{DBLP:conf/kdd/WangC016}, \cite{DBLP:conf/emnlp/WestonCA14}. Furthermore, modifying the properties of random walks can assist the learned embeddings in encapsulating both local and global graph properties \cite{DBLP:conf/cikm/CaoLX15}, \cite{DBLP:conf/asunam/PerozziKCS17}, \cite{DBLP:conf/kdd/WangC016}. The problem of {\it heterogeneous} graph embedding was addressed with metapath2vec \cite{Dong:2017:MSR:3097983.3098036}, where metapaths among specific entities types are defined and then random walks are generated only in accordance to those metapath schemes. This approach was extended in \cite{cen2019representation} to include node attributes and multiplex edges. Further advancements have allowed the incorporation of node and edge feature vectors ({\it auxiliary information}) to facilitate inductive learning of representations \cite{DBLP:journals/corr/HamiltonYL17}. In these works, the estimation of node embeddings proceeds through typical sampling-based approaches \cite{gutmann10a, Gutmann2012}.

\subsection{Large Scale Applications In Industry}
Many internet-scale recommendation systems use graph embedding techniques to supply millions of customers with potentially useful or interesting content related to their past interests \cite{DBLP:conf/kdd/GrbovicC18}, \cite{lerer2019pytorchbiggraph}, \cite{DBLP:conf/kdd/WangHZZZL18}, \cite{DBLP:conf/kdd/YingHCEHL18}. These systems typically model vertices as users, content, or products on very large graphs, and several instances of graph embeddings techniques \cite{gao2018large} have been applied to networks with millions of unique entities, with even a few applications in the financial services space \cite{DBLP:journals/corr/abs-1802-04198} using autoencoders to create embeddings on account transaction data. Embeddings from these approaches are typically used in downstream applications like product recommendation \cite{DBLP:conf/kdd/WangHZZZL18}, \cite{ying2018graph} and maximizing proper ad placement \cite{ordentlich2016network}. This transfer learning approach is very similar to the impact that word embeddings have had for a variety of NLP tasks \cite{DBLP:journals/corr/MikolovSCCD13}, \cite{DBLP:conf/naacl/DevlinCLT19},   \cite{DBLP:journals/corr/abs-1906-08237}. To our knowledge, the method proposed in this paper is the first application of graph-embeddings to financial transactions.

\section{Methodology}
\label{methodology}

\begin{figure*}
\includegraphics[width=\textwidth, keepaspectratio]{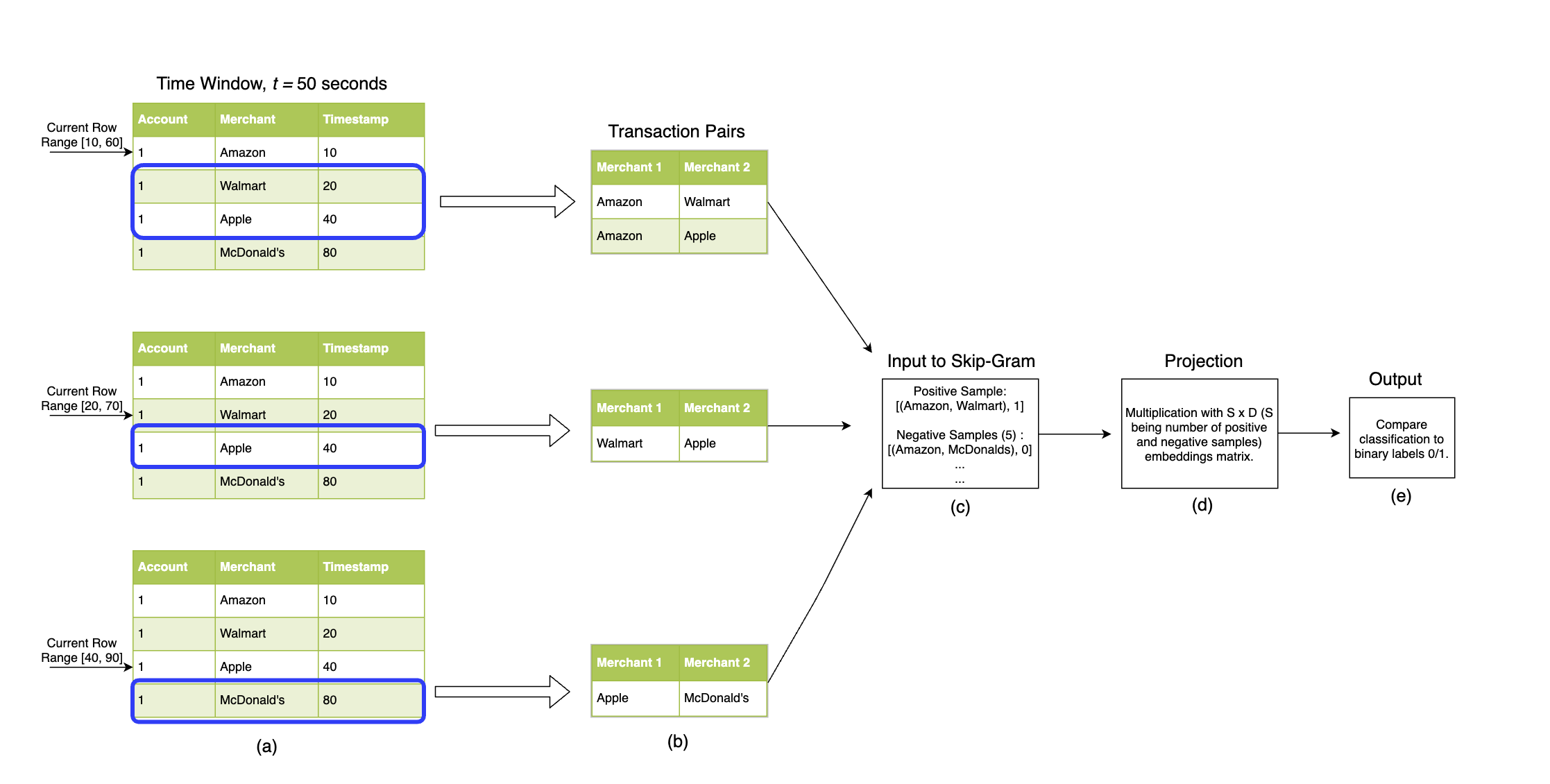}
\caption{Model pipeline, from data pre-processing to training with Skip-gram. Using stringent time windows and pairs of transaction pairs (example time window, \textit{t} = 50 seconds) allows for creating meaningful embeddings on graphs with millions of unique entities. }
\label{model_pipeline}
\end{figure*}

\subsection{Projections of Heterogeneous Graphs}
The credit card transaction graph is a bipartite graph between accounts and merchants, with a transaction forming an edge. While much previous work has studied embeddings of homogeneous graphs, \cite{Dong:2017:MSR:3097983.3098036} consider heterogeneous graphs in their \emph{metapath2vec} framework. Here, a metapath scheme is chosen to determine which sequence of node types are considered in the walks. Then, only random walks that are consistent with this scheme are generated for training embeddings. Concretely, in our bipartite graph we might consider a metapath such as \{\emph{Account}, \emph{Merchant}, \emph{Account}\} or \{\emph{Merchant}, \emph{Account}, \emph{Merchant}\}. Given these schemes, we would only consider graph walks that adhere to such triplets: identifying accounts as similar because they shop at the same merchant (and vice versa). Alternatively, we can reach similar training sets by instead inducing two \emph{homogeneous} projection graphs derived from the original bipartite graph: an accounts graph and a merchants graph. Short random walks on these homogeneous graphs will induce the same training pairs as those that would have been generated according to the short metapath schemes. As described below, this approach brings enhanced computational gains and flexibility. The results and analyses reported here will focus exclusively on the merchant embeddings.

\subsection{Pre-processing Stage: Modeling a Graph as Pairs of Transactions}
\label{methodology:preprocessing}
In the projection graph(s), an edge between two merchant vertices represents the presence of at least one account who made transactions at both merchants within a specified time window. The more often an account shops at the same two merchants within a fixed time window, the greater the weight on that edge. We read all transactions into a table that fits in memory, as shown in part (a) of Figure \ref{model_pipeline}. We then transform this table into pairs of transactions by only keeping merchants that processed a transaction from the same account-holder within a specified time window. We set the time window depending on the node type that we are looking at as well as the density of the graph. For instance we would want a smaller window for grouping similar accounts on a merchant than we would for grouping merchants on an account. In general, we find that increasing the time window too large can lead to significantly more connections between unrelated merchants, which decreases the quality of the embeddings. 

Storing this information within a table allows the user to distribute the aforementioned time windowing strategy over all rows of merchants, instead of individually running random walk operations from every merchant vertex \footnote{While algorithms for distributed random walks on graphs do exist, these techniques are non-trivial to implement and excessive for creating pairs of transactions (or equivalently, short walks of length 2).}. Part (b) in Figure \ref{model_pipeline} shows examples of transaction pairs after time-windowing. Multiple transaction pairs with the same two merchants indirectly represent a weighted edge between those merchants on a graph. In effect, this formulation allows us to model a weighted graph without explicitly creating one.

\subsubsection{Brand Level Merchant Names vs Raw Merchant Names}
When creating training pairs, we consider two different approaches. The first is using the raw merchant name and appending it with the zip code. The raw merchant name differentiates between franchise locations, but some unrelated merchants have the same raw merchant name. For this reason, the zip code is appended to the name to properly differentiate. This gives us many  merchants to work with, as well as a less dense graph as accounts are less likely to be paired together. Table \ref{table:datasets} shows that the raw merchant graph contains approximately $10^6$ merchants and $10^9$ edges. 

The other approach is to use the brand-level merchant name which rolls up all franchises to the same name. This creates a highly interconnected graph, as everyone nationwide who shops at a particular brand is likely to be paired with another similar shopper. This causes the number of pairs formed with the brand-level graph to be far larger than those of the raw merchant. Additionally, for rarely-occurring merchants, it can be challenging to accurately identify the correct brand entity. Due to these factors, we drop any merchants with fewer than 50 transactions per day. Table \ref{table:datasets} shows that the brand-level graph contains approximately $10^4$ merchants and $10^7$ edges.

\begin{table}
\begin{center}
\begin{tabular}{c|cc|}
\cline{2-3}
\textbf{}                      & \multicolumn{1}{l}{\textbf{Brand-Level Graph}}   & \multicolumn{1}{l|}{\textbf{Raw Merchant Graph}} \\ \hline
\multicolumn{1}{|c|}{\# nodes} & \multicolumn{1}{c|}{$\sim10^4$} & $\sim10^6$ raw merchants        \\
\multicolumn{1}{|c|}{\# edges} & \multicolumn{1}{c|}{$\sim10^7$} & $\sim10^9$                      \\ \hline
\end{tabular}
\end{center}
\caption{Dataset descriptions. For the brand-level and raw merchant graphs. Note that the larger raw merchant dataset more closely reflects the size of a typical transaction graph.}
\label{table:datasets}
\end{table}

\subsubsection{Separating Online from Offline Transaction Pairs}
Prior to training, we place online merchants and physical merchants into two distinct tables. When looking at raw merchant names we find that online retailers - e.g \textit{Amazon.com}, \textit{Newegg.com} - often precede or follow several other unrelated merchants in transaction sequences. Even though there are physical merchants that also frequently appear in many transaction pairs, these vendors are typically separated into several stores with distinguishable identifiers - e.g \textit{McDonalds 94123}. Consequently, each store is likely to be close to other merchants in the same geographic region.  Online merchants, however, are location-agnostic and similar to supernodes on a graph that have a disproportionately high number of edges. Training on each set separately allows us to create embedding representations for merchants that are not biased by artificial relationships in the raw transactional data.

\subsection{Approximating DeepWalk with Transaction Pairs}
We posit that for this financial graph, short-range interactions (one-hop and two-hop neighborhoods) will be sufficient to yield effective embeddings. When viewed in this way, we can consider random walks on a graph in the limit of either (i) very short walk lengths or (ii) very short context windows within walks. At this short-range limit, we consider only two-hop walks on the bipartite graph, or equivalently, one-hop walks on the homogeneous projection.  Our embeddings are trained using negative sampling, with node-pairs generated as just described used as positive samples. Negative samples are generated by sampling nodes at random (assuring no actual edge exists),  using the negative sampling strategy described in \cite{DBLP:journals/corr/abs-1301-3781}.  This leads to optimization of the following loss:

% Training on pairs effectively reduces the Skip-Gram algorithm to maximize the co-occurrence probability of merchants that frequently appear with one another in the training set of transaction pairs. By constraining the sequence length, we achieve a tighter optimization objective: 

%     \begin{equation} \label{eq:trueobjectivefunc}
% \displaystyle{\minimize_{\phi}} -log~P(\phi(m_{i+1}) | \phi(m_{i}))
%     \end{equation}

% where $m \in V$ denotes a merchant selected from a set of unique merchant entities, $V$. The mapping function $\phi: m \in V \mapsto \R^{d} $ retrieves the embedding representation for any given merchant. Given that there can be up to millions of unique merchants, directly calculating equation (\ref{eq:trueobjectivefunc}) is not feasible. We instead use Skip-Gram with negative sampling (SGNS) to efficiently train this system. We present a visual description of this step in part (c) of Figure \ref{model_pipeline}, and a more manageable objective function below:

\begin{multline}
\label{eq:approximateobjectivefunc}
\displaystyle{\minimize_{\mathcal{\phi}}  \mathcal{L}(\phi)} = -\sum_{m \in V}[log~P(y=1 | \phi(m), \phi(m_{pos})) \\ +  kE[log~P(y=0 | \phi(m), \phi(m_{neg})]],
\end{multline}

where $m \in V$ denotes a merchant selected from a set of unique merchant entities, $V$, the mapping function $\phi: m \in V \mapsto \R^{d} $ retrieves the embedding representation for any given merchant, $m_{pos}$ denotes a positive sample merchant for a given merchant, and $m_{neg}$ denotes a negative sample chosen from $V$ and $k$ is the number of negative samples. 

If we interpret equation (\ref{eq:approximateobjectivefunc}) from a graph-based perspective, minimizing this objective function amounts to creating embedding representations that capture first-order proximity relationships between different merchants on a graph. Representing the transactional graph as pairs forces the model to capture these first-order relationships. Part of the motivation for not exploring higher-order relationships is to guarantee true noise samples during training. As transactional data is often noisy, maintaining a small context window makes it easier to guarantee that a negatively sampled merchant does not appear in the immediate vicinity of a merchant of interest. A larger context window also increases the probability of interrelating merchants that are not actually meaningfully similar. However, the use of the time-window strategy does allow for a tune-able parameter that can act similarly to capturing higher order relationships. Intuitively, if positive pairs are generated from all merchants that a single account has shopped at in an entire month, then this is a higher order proximity than if only a one hour time window had been considered. 

Furthermore, \cite{zheng2013comparison} demonstrates that if the number of random walks per vertex is large enough, the expected average walk length for each one of them will converge to the shortest path between source and destination vertex. In other words, by significantly increasing the number of walks, the expected walk length reduces to that of the shortest path. Thus, using transaction pairs - or truncated random walks of length 2 - serves as an approximation to the shortest path between a merchant and every semantically similar merchant. Our qualitative and quantitative analysis in Section \ref{results} demonstrates that this approximation is effective.

\section{Results}
\label{results} 

We train our model on both merchant datasets for roughly 2 epochs on 48 vCPUs. For quantitative analysis, we present F1, link prediction AUC and area under the precision-recall curve for evaluating our embeddings with an internal fraud detection tool. We report F1 and LPA scores from running experiments on both the raw and brand-level merchant datasets, but use the brand-level dataset for presenting embeddings visualizations due to its smaller size.

\begin{figure}
\centering
\includegraphics[width=7cm, keepaspectratio,]{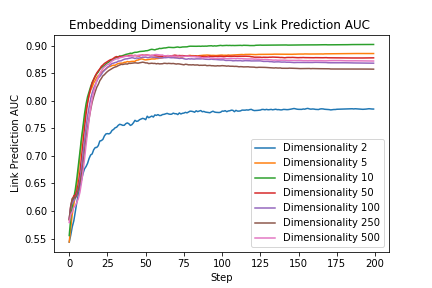}
\caption{Impact of embedding dimensionality. While increasing embedding dimensionality yields very large quantitative improvements on LPA at low dimensionality, results quickly stagnate after 10 or more dimensions and overfitting is observed. }
\label{fig:dim_fig}
\end{figure}

\begin{table}
\begin{center}
\resizebox{9cm}{!}{
\begin{tabular}{l|cc}
 & \multicolumn{1}{c}{\textbf{Link Prediction AUC (LPA)}} & \multicolumn{1}{c}{\textbf{F1 Score}} \\ \hline
Brand-Level Merchant Graph & 0.90 & 0.67 \\ 
Raw Merchant Graph & 0.68 & 0.67 \\

\end{tabular}}

\end{center} 
\caption{Model Performance after 2 epochs of training on all datasets. The model likely performs very well on the smaller branded dataset due to fewer instances of overrepresented / underrepresented merchants in this graph. }
\label{table:metrics}
\end{table}

\subsection{Performance on Brand-Level Network} 
\subsubsection{Quantitative Analysis}
We start by analyzing our model's effectiveness on the brand-level graph. In Table \ref{table:metrics}, we see that the model achieves a link prediction AUC of roughly 0.91 and a F1 score of 0.67. The model's high performance on both metrics demonstrate its capability to learn meaningful relationships even within smaller-sized graphs of roughly 10,000 or fewer elements. This performance also demonstrates that training on pairs of transactions (as opposed to long sequences) is not detrimental to creating high-quality embeddings. This is to be expected, as training Skip-gram with a sequence length of 2 simplifies its objective function to optimize  at every iteration. This in turn forces the system to create node embeddings that encode meaningful information about the edges in their surrounding subgraph.

In Figure \ref{fig:dim_fig}, we present an analysis on how embedding dimensionality affects the model's performance. We see that increasing dimensionality from 2 to 5 yields a roughly 13\% improvement in link prediction AUC, but further increasing this hyperparameter does not yield significant benefits on this particular metric. This is likely due to the brand-level dataset's small vocabulary size (roughly $10^4$ unique merchants). This diagram suggests that only a small embedding dimensionality is required to adequately capture semantic similarity for the brand-level transactional dataset discussed in this paper.

\subsubsection{Qualitative Analysis}

\begin{table}
\centering
\begin{tabular}{lll}
\begin{tabular}{c}
 \textbf{Adidas} \\ \hline \hline
 Nike    \\      
 Reebok    \\      
 Timberland  \\       
\end{tabular}
\hspace{8pt}
\begin{tabular}{c}
 \textbf{Delta Air Lines}  \\ \hline \hline
 United Airlines    \\       
 American Airlines    \\      
 Frontier Airlines      \\   
\end{tabular}
\hspace{8pt}
\begin{tabular}{c}
 \textbf{West Elm}  \\ \hline \hline
 Anthropologie       \\   
 Crate\&Barrel         \\ 
 Pottery Barn         \\
\end{tabular}
\end{tabular}
\\
\begin{tabular}{lll}
\begin{tabular}{c}
 \textbf{KFC}  \\ \hline \hline
 Taco Bell      \\    
 Little Caesars   \\       
 Burger King        \\ 
\end{tabular}
\hspace{8pt}
\begin{tabular}{c}
 \textbf{Starbucks}  \\ \hline \hline
 Panera Bread         \\ 
 Dunkin Donuts          \\
 Jamba Juice         \\
\end{tabular}
\hspace{8pt}
\begin{tabular}{c}
 \textbf{Gap}  \\ \hline \hline
 Banana Republic         \\ 
 Ann Taylor Loft          \\
 J Crew         \\
\end{tabular}
\end{tabular}
\caption{Merchant nearest neighbors derived from the brand-level merchant graph. For each query merchant vector (in bold), the top three nearest neighbor merchant vectors are shown.}
\label{neighbors}
\end{table}

The brand-level merchant graph embeddings provide intuitive consistency. With word embeddings, a common observation is that words which are semantically similar tend to be embedded in close proximity. Here, we redefine semantic similarity to be sets of merchants which are interchangeable for any given commerce purpose. That is, two merchants are semantically similar if they exist in the same industry, category, price point or all of the above.  With this in mind, we report in Table \ref{neighbors} the nearest neighbor merchant vectors for several household brands. For example, we see that the nearest neighbors to the \emph{KFC} vector are not only other restaurants, but other fast food competitors. As can be seen in Table \ref{neighbors}, this holds true across many industries including fashion, air travel, food, and furniture. The \emph{West Elm} vector presents an interesting result. Two of the three nearest neighbors are obviously correct: \emph{Pottery Barn} and \emph{Crate\&Barrel} are also furniture manufacturers. However, the closest neighbor in the entire merchant space is actually \emph{Anthropologie}, a store which is most commonly known as a fashion brand. However, within \emph{Anthropologie's} offerings is an extensive home furnishings and furniture section. Overall, these findings indicate that semantically similar merchants are typically embedded close together.

\begin{figure*}
\includegraphics[width=17.5cm, keepaspectratio,]{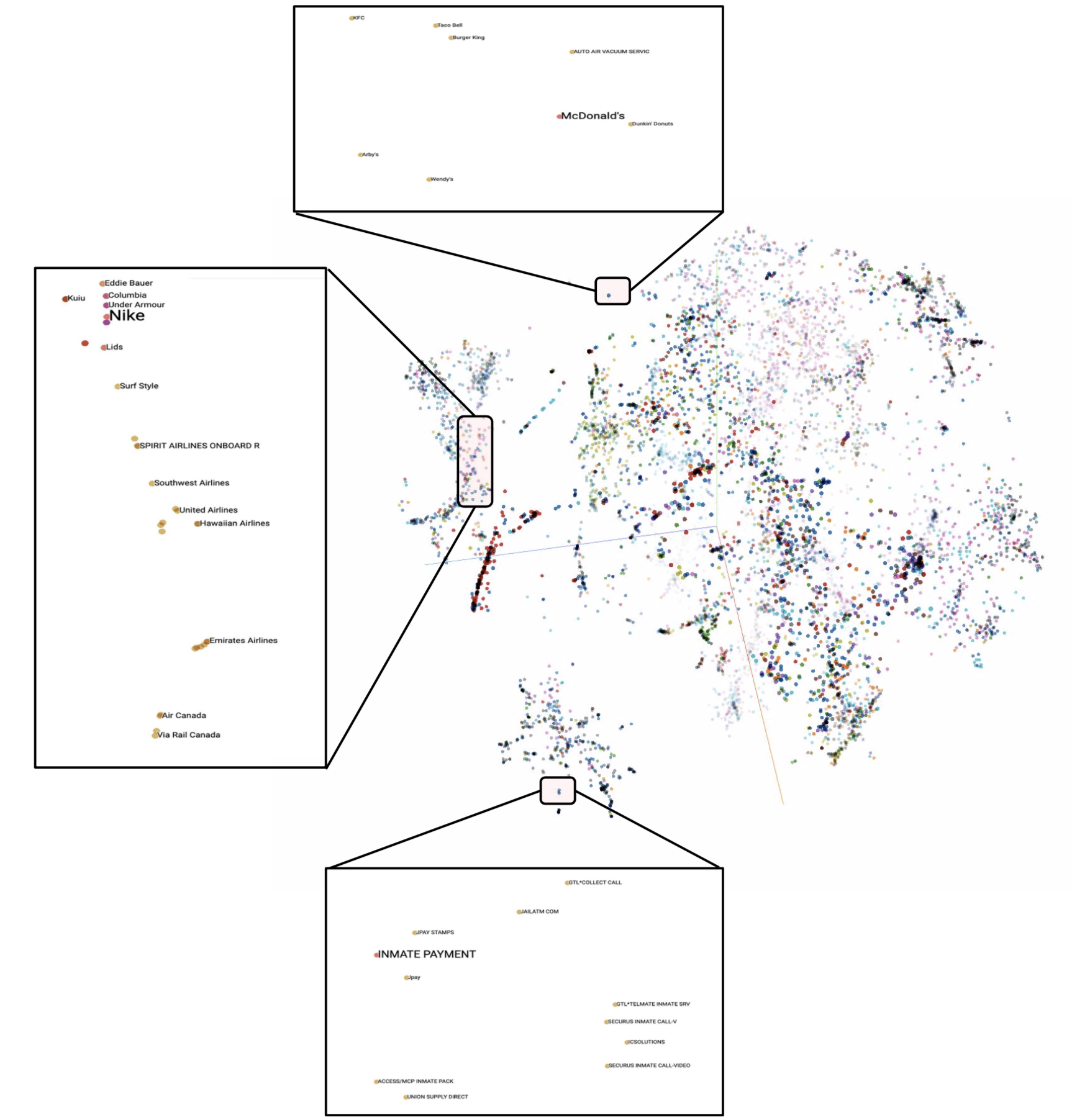}
\caption{Example clusters found in a two dimensional t-SNE projection of brand-level embeddings. Best viewed digitally.}
\label{tsne_fig}
\end{figure*}

This general pattern of separability-by-industry extends across the whole embedding space. Figure \ref{tsne_fig} shows a low-dimensional visualization of the embedding space for our brand-level merchant graph. As expected, merchants which serve the same industry or category tend to co-locate in similar areas of the embedding space: a visual extension of the results of Table \ref{neighbors}. In the left of Figure \ref{tsne_fig}, we can note that sporting goods brands such as \emph{Nike}, \emph{Under Armour}, and \emph{Columbia} are located close to each other. Similarly, note that airlines are co-located, as well as fast food (top). Finally, the bottom close-up shows a region of the embedding space with merchants such as \emph{Jpay}, \emph{INMATE PAYMENT}, \emph{SECURUS INMATE CALL-V}. These companies provide a set of services for telecommunications and payments into and out-of the American prison system. That is, the customers of these companies are inmates (and family members of inmates) who must use these merchants to conduct common activities. Due to these specialized services, there should be very little overlap between these merchants and any accounts not affiliated with inmates. It is encouraging that our graph embedding system is able to accurately embed these merchants together. 

A final set of questions we can ask with these brand-level merchant embeddings is whether relationships between merchants can be consistently observed within and between industries. In much the same way that analogical reasoning tasks can be accomplished with word vectors, it might be feasible to identify compelling relationships between merchant vectors. 

One way to construct analogies for merchants is to devise relations between merchants within one category and determine if the same relationship holds (geometrically) across categories. As a concrete example, we can recognize that within a particular industry, there will exist several offerings that are not direct competitors but instead operate at different price points. For example, within automobiles, the typical \emph{Porsche} price point is well above that of \emph{Honda}, though they offer the same basic good. If there is a direction in the embedding space that encodes price point (or \emph{quality} or \emph{luxury}) then this component should contribute to all merchants across all industries. In this way, the embeddings can elucidate analogies such as "Porsche is to Honda as \emph{blank} is to Old Navy". 

\begin{figure}
\centering
\includegraphics[width=7cm, keepaspectratio,]{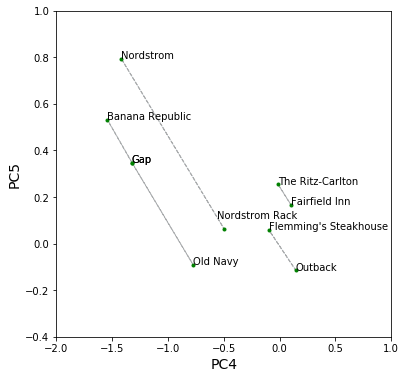}
\caption{Merchant embeddings tend to encode typical price point of goods.  Merchants from multiple industries visualized in the subspace spanned by the fourth and fifth principal components. Pairs of merchants are joined by dotted lines, with each pair containing a high-end merchant and a more affordable counterpart. Across disparate industries, the relationship between high-end offerings and affordable offerings is embedded in a consistent direction of this space.}
\label{fig:price_analogies_fig}
\end{figure}

Uncovering such a direction can be achieved in several ways (\cite{DBLP:conf/conll/LevyG14}, \cite{DBLP:conf/naacl/MikolovYZ13}). For our purposes, we assumed that such a direction, if it existed, was likely a dominant source of variation between the embeddings and that this direction is likely captured by one or a few principal components of variation. With this in mind, we identified several pairs of merchants which exist in the same category yet span a range of price points. These included: \{\emph{Gap} and \emph{Old Navy}\}, \{\emph{Nordstrom} and \emph{Nordstrom Rack}\}, \{\emph{Ritz-Carlton} and \emph{Fairfield Inn}\}, and so on. These pairs were chosen because they represent nearly identical offerings within a category, but at a high and low price point. In Figure \ref{fig:price_analogies_fig}, we visualize these paired merchant vectors as projected into a convenient subspace spanned by two principal components. Interestingly, the relationship (slope) between the low-end and high-end offering is nearly parallel for all of these pairs. There indeed exists a \emph{direction} in the embedding space which encodes price point and this direction is consistent across these disparate industries. 

\subsection{Performance on Raw Merchant Network}

\begin{table}[h]
\begin{center}
\resizebox{8.2cm}{!}{\begin{tabular}{p{4cm}|p{2cm}}
 & \multicolumn{1}{c}{\textbf{\%$\Delta$ in AUpr}} \\ \hline
Fraud Detection Tool + \\ Raw Merchant Embeddings & +0.9\% 
\\
Fraud Detection Tool + \\ MLP[Raw Merchant Embeddings] & +5.2\%
\end{tabular}}

\end{center} 
\caption{Training an internal fraud detection tool with only merchant embeddings or with projected merchant embeddings (through a multi-layer perceptron) yields higher classification performance over an internal baseline model.}
\label{table:fraudmetric}
\end{table}

\subsubsection{Quantitative Analysis}

We see in Table \ref{table:metrics} that the model performs similarly on the raw merchant graph with respect to F1 score. Achieving a 0.67 score on the one million plus merchant graph demonstrates the model's ability to maintain meaningful embedding representations that are not heavily influenced by outliers in the larger-sized graph. While the model scores lower on lower link prediction AUC (0.68) for this dataset, we expect to see a drop-off due to training on a sparser graph; there are far more instances of merchants in this network that are only connected to a few neighboring vendors. Still, even with this obstacle, the model is able to create embedding vectors that are semantically meaningful as indicated by the top-5 closest neighbors to the examples given in Table \ref{table:rawmerchant}. For example, the first two columns show fast-food merchants that not cluster together by type - e.g \textit{Dunkin}, \textit{Nature's Way Cafe} - but also by geographic region. 

\subsubsection{Qualitative Analysis}
Table \ref{table:rawmerchant} shows nearest neighbors for several raw merchant vectors. Since these raw merchant entities correspond to physical locations, it is not surprising that geography plays a large role in this embedding space. In Table \ref{table:rawmerchant}, each entity is shown alongside its zipcode, confirming that merchant vectors tend to be embedded near other merchants in the same geographic area. (Note that zipcode and geography are not inputs to the model). We see that the top-5 neighbors for \textit{DUNKIN \#332240 Q35}  are not only other stores in the same chain - e.g \textit{DUNKIN \#341663}, \textit{DUNKIN \#341663 Q35} - but even loosely related cafes like \textit{NATURE'S WAY CAFE BO} that lie within neighboring counties in Florida. Further, the nearest neighbors here highlight what is potentially a naming or logging error in the point-of-sale system. Note that the same physical store location \emph{DUNKIN \#341663} shows up under two entity names: \emph{DUNKIN \#341663 Q} and  \emph{DUNKIN \#341663 Q35}. This demonstrates an interesting potential for this kind of analysis to be leveraged for entity resolution based not on string similarity, but based on correlated shoppers. Finally, the second and third examples illustrate some ways that the model can capture local geographic cultural nuances such as the high proportion of breweries in Portland, Oregon where Powell's Burnside is located, when compared the number of coffee shops in Los Angeles, CA where The Last Book Store is located.

    \begin{table}[h]
        \begin{center}
                \tiny
                \resizebox{9.2cm}{!}{\begin{tabular}{|l l|l l|l l|}
         \textbf{DUNKIN \#332240 Q35} & \textbf{33442} & \textbf{POWELL'S BURNSIDE} & \textbf{97209} & \textbf{THE LAST BOOK STORE} & \textbf{90013} \\
         CHUCK E CHEESE  682                    & 33428          & TRIMET TVM                        & 97202          & PAMPAS GRILL- STYL           & 90036          \\
         DUNKIN \#341663       Q        & 33442          &  TARGET 00027912                   & 97205          &SQ *BLUE BOTTLE COF          & 90013          \\
        DUNKIN \#341663       Q35                         & 33442          & 10 BARREL BREWING CO              & 97209          &  HIGHLAND PARK BOWL           & 90042          \\
         NATURE'S WAY CAFE BO        & 33431          & DESCHUTES BREWERY                & 97209          & VERVE COFFEE ROASTERS        & 90014          \\
        DD/BR \#338392       Q3              & 33073          &  PORTLAND JAPANESE GARD          & 97205          & GRILL CONCEPTS - S           & 90404         
        \end{tabular}}
        \end{center}
        \caption{Examples of merchant similarities when trained using the raw merchant name. Each merchant is shown alongside its zip code. Geography is a strong signal in these embeddings, but semantic and regional influence can also be observed.}
        \label{table:rawmerchant}
        \end{table}

\subsection{Application to Fraud Detection}

We also assess the quality of our raw merchant embeddings by evalauting them in a transfer learning task involving transaction fraud detection. Results on these experiments are reported relative to a baseline model (the details of which we omit here) and are quantified by the area under the precision-recall curve (AUpr). In Table \ref{table:fraudmetric}, we see that directly using the trained embeddings from the raw merchant graph yields a roughly 1.0\% improvement in fraud classification AUpr when these embeddings are included as additional features to the model. One downside of this approach is the large expansion of the feature space in order to include the embeddings. To overcome this, we additionally tested a model whereby we added a trainable MLP that took as input the embedding for a transaction's merchant and predicted a binary output for fraud. This ancillary model, once trained, can then be used to output a single fraud score per transaction, conveying only the information contained in the merchant embedding space that is useful for fraud detection. This single score is then passed into the base model as an additional feature and yields a 5.2\% boost in efficacy. Merchants that engage in fraudulent transactions typically engage with similar vendors over all transaction pairs. As our model's objective is to encode transactional behavior within each merchant's embedding representation, we attribute an increase in classification accuracy to capturing semantically meaningful features that allow the classifier to better identify likely-fraudulent merchants.

\section{Conclusion}

In this paper, we propose an approach for training embeddings of entities from financial transactions. Our approach poses sequences of financial transactions as a graph, where a customer engaging in a transaction at two merchants within a specified time window constitutes an edge between those two merchant in the network. We demonstrate that this approach results in semantically meaningful embedding vectors for up to millions of unique merchant entities. We quantitatively show in Section \ref{results} that our model scores strongly with respect to link prediction AUC and F1 evaluation scores, and also provides lift in classification accuracy for an internal fraud detection tool. 

The results presented here were primarily based on capturing network topology only, while omitting ancillary attributes about accounts, merchants, and transactions. Future work remains to be done to incorporate techniques such as those described in (\cite{DBLP:journals/corr/GroverL16}, \cite{cen2019representation}). Lastly, we hope to analyze how embeddings impact other downstream applications in the financial services such as marketing and credit charge-off prediction.

\bibliography{deeptrax}
\bibliographystyle{deeptrax}

\end{document}